\pdfoutput=1

\documentclass[10pt,twocolumn,letterpaper]{article}
\usepackage{cvpr}
\usepackage{times}
\usepackage{epsfig}
\usepackage{graphicx}
\usepackage{amsmath}
\usepackage{amssymb}

\graphicspath{{img/}} 
\usepackage{bm}
\usepackage{epigraph}

\let\originalepigraph\epigraph 
\renewcommand\epigraph[2]{\originalepigraph{\textit{#1}}{\textsc{#2}}}
\usepackage{dsfont}
\usepackage{subcaption}
\newtheorem{definition}{Concept}[section]
\usepackage{txfonts}
\usepackage[dvipsnames]{xcolor}

\usepackage[breaklinks=true,bookmarks=false]{hyperref}

\cvprfinalcopy 

\def\cvprPaperID{****} 

\begin{document}

\title{Towards an Understanding of Neural Networks in Natural-Image Spaces}

\author{Yifei Fan \qquad Anthony Yezzi\\
Georgia Institute of Technology\\
85 5th Street, NW Atlanta, GA 30308 USA\\
{\tt\small yifei@gatech.edu, anthony.yezzi@ece.gatech.edu}
}

\maketitle

\begin{abstract}
   Two major uncertainties, dataset bias and adversarial examples, prevail in state-of-the-art AI algorithms with deep neural networks. In this paper, we present an intuitive explanation for these issues as well as an interpretation of the performance of deep networks in a natural-image space. The explanation consists of two parts: the philosophy of neural networks and a hypothetical model of natural-image spaces. Following the explanation, we 1) demonstrate that the values of training samples differ, 2) provide incremental boost to the accuracy of a CIFAR-10 classifier by introducing an additional ``random-noise'' category during training, 3) alleviate over-fitting thereby enhancing the robustness against adversarial examples by detecting and excluding illusive training samples that are consistently misclassified. Our overall contribution is therefore twofold. First, while most existing algorithms treat data equally and have a strong appetite for more data, we demonstrate in contrast that an individual datum can sometimes have disproportionate and counterproductive influence and that it is not always better to train neural networks with more data. Next, we consider more thoughtful strategies by taking into account the geometric and topological properties of natural-image spaces to which deep networks are applied.
\end{abstract}

\section{Introduction} \label{sec:intro}
Recent years have witnessed the rapid development of artificial intelligence (AI) and deep learning. Deep networks distinguish themselves in various types of computer-vision challenges such as ImageNet \cite{deng2009imagenet} and COCO \cite{lin2014microsoft}, and some AI algorithms even outperform humans on certain tasks and test sets \cite{stallkamp2012man}, \cite{schroff2015facenet}. 
However, two major issues remain and prevent us from establishing robust real-world applications with current algorithms. One is dataset bias \cite{torralba2011unbiased}, meaning that a machine-learning algorithm that performs well on one dataset may fail on another. The other is adversarial examples \cite{szegedy2013intriguing}, which shows that tiny modifications on inputs may lead to incorrect outputs by deep networks, even though the perturbations are almost imperceptible by humans. 

In this paper, we present our understanding of neural networks in natural-image spaces and
a model of natural-image spaces which manage to explain the uncertainties. 
The contributions of the paper are as follows:  
\begin{itemize}
\item We provide a unified explanation for dataset bias and adversarial examples, the two prevailing uncertainties in neural networks, from the perspective of variational calculus and properties of natural-image spaces.  
\item We illustrate that the values of training samples differ. Training with more samples does not guarantee higher accuracy, and even random noise can sometimes help improve the performance of neural-network classifiers. 
\item We present a hypothetical model for natural-image spaces. The model can potentially guide a network to alleviate over-fitting and enhance robustness against adversarial examples, as well as improve accuracy. 
\end{itemize}

The rest of the paper is organized as follows. Section \ref{sec:literature} reviews related studies. Section \ref{sec:philosophy} illustrates the philosophy of neural networks,  explains in theory why the training process is data-dependent and zero-constrained, and specifies the flaw arising from limited options in data labels. Section \ref{sec:image_space} examines the characteristics of image data and presents the hypothetical image-space model. Sections \ref{sec:experiments} and \ref{sec:discussion} follow with experiments and discussion, respectively. 
All relevant materials that accompany this paper will subsequently be available on
\href{https://github.com/thelittlekid/natural-image-spaces}{GitHub}.

\section{Related Work} \label{sec:literature}
\emph{\bf{Dataset bias}}: The dataset bias issue was first pointed out in \cite{torralba2011unbiased}, after the authors became aware that researchers who had worked in object and scene recognition could easily guess which images came from which dataset. They claimed that such a difference among datasets could even be captured by a classifier and reflected in the diagonal pattern of the confusion matrix. Although the purpose of computer vision datasets is to offer unbiased representations of the world, they seem to express them with a strong built-in bias. A follow-up study \cite{khosla2012undoing} exploited dataset bias during the training phase. The authors proposed two sets of weights, bias vectors and visual world weights to undo the damage of dataset bias. They demonstrated the benefit of explicitly accounting for bias when multiple datasets are involved. Another study \cite{tommasi2014testbed} provided a large-scale analysis of the issue with twelve databases. 
However, as most previous studies have addressed the issue on dataset level, they do not help explain or eliminate such a bias in practice. It is possible that the data samples we collect resemble multiple datasets at the same time. Thus, stating the characteristics of images that lead to such bias is beneficial. One apparent bias stems from the size of targets. The evaluation metric of COCO \cite{lin2014microsoft} challenges, which emphasizes the effects of iconic and non-iconic objects, accounts for the size factor.

\emph{\bf{Adversarial examples}}: As an intriguing property first discovered by \cite{szegedy2013intriguing}, a hardly-perceptible perturbation on inputs could lead to misclassification. Since then, researchers have been working on methods of attacking/fooling or defending deep networks. 
On the attackers' side, the authors of \cite{nguyen2015deep} forced the network to generate high-confidence predictions for images that are even unrecognizable by humans. 
With DeepFool \cite{moosavi2016deepfool}, we can easily change the output label for an image with unnoticeable perturbations. Furthermore, we can fool the network with very few pixels \cite{su2017one}. 
In general, two types of perturbations exist: adversarial and universal \cite{moosavi2017analysis}. While the former depends on individual image, the latter is a uniform perturbation applied to all images. The existence of universal perturbation implies important characteristics of the geometry of the decision boundaries in high-dimensional space \cite{moosavi2016universal}. In \cite{mopuri2017fast}, universal perturbations are computed without any knowledge of the target data.  
On the defenders' side, efficient defenses against adversarial attacks are provided in \cite{zantedeschi2017efficient}. The idea is to stabilize the prediction by reinforcing the weak points of the network. A recent study \cite{madry2017towards} proposed methods for training networks with improved resistance to adversarial attacks.
In practice, studies do not agree on the extent of the effects of these attacks. Perturbations for object detection in autonomous vehicles is not a concern \cite{lu2017no}. New attacks, however, generate more subtle adversarial examples \cite{evtimov2017robust}. 
Nevertheless, the combat between attackers and defenders will guide us to more robust and secure networks.  

\emph{\bf{Image manifold}}: The idea of the image manifold comes from smooth manifold in mathematics \cite{lee2003smooth}. 
A topological space $M$ is a topological $n$-manifold if it is 1) a Hausdorff space, 2) second countable, and 3) locally Euclidean of dimension $n$. Manifold $M$ has important topological properties, including the following: 1) It is connected if and only if it is path connected, and 2) it is locally compact. 
It is assumed that natural images are embedded in a manifold whose dimension is much lower than that of the images themselves. 
Such a formulation consists of two key components: a distance metric and a map from parameter (or feature, latent) space to the image space. 
In computer vision, researchers manage to approximate the image manifold via learning \cite{he2004learning}. The generative adversarial network (GAN) is used in the state-of-the-art approximation of the image manifold \cite{zhu2016generative}. However, such approximation may not fully account for that spaces of images are vast and sparsely populated \cite{hassabis2017neuroscience}.

\section{Philosophy of Neural Networks} \label{sec:philosophy}
In this section, we analyze the high-level in machine learning from the view of variational calculus and reveal potential reasons for uncertainties in neural networks. 
\subsection{The Variational Perspective on Learning}
The learning process is generally based on the idea of gradient descent \cite{cauchy1847methode}: If $f(\cdot)$ is defined and differentiable in the neighborhood of point $a$, then $f(x)$ decreases fastest if one goes from $a$ in the direction of the negative gradient of $f$ at $a$ (i.e., $-\nabla f(a)$). In other words, if $a^{n+1} = a^n - \gamma\nabla f(a^n)$ for small enough $\gamma$, then $f(a^n)\geq f(a^{n+1})$. As a result, gradient descent algorithms are likely to find an extremum point of $f(\cdot)$.  
However, one crucial difference exists between the function $f(\cdot)$ and the underlying target function $h(\cdot)$ to be learned by neural networks. Here $f(\cdot)$ is the objective function that we maximize/minimize. After iterations, we obtain an extremum point $a^*$ and its corresponding extremum $f(a^*)$. In contrast, according to the geometric view taken in functional analysis, the \textbf{target function $h(\cdot)$ itself is an extremum point} similar to $a^*$, whose values may be explained as components of an infinite-dimensional vector indexed\footnote{E.g., the index set of $v = (v_1, v_2, v_3)$ is $\{ 1, 2, 3 \} $} by its domain $X$, that is, $(h_x)_{x\in X}$. 
On the contrary, the \textbf{objective/loss functions in machine learning correspond to functionals $L(h(\cdot))$ in calculus of variations}. 

A functional is a special type of function that takes another function as its input. A simple example of a functional is as follows.  
\begin{equation} \label{eq:calculus_of_variations_continuous}
L(f, x) = \int_0^1{f(x) - y(x)} \, dx,  \quad y(\cdot)\,\text{is given}
\end{equation}
We can easily discretize a functional and obtain a loss function for neural networks. 
\begin{equation} \label{eq:calculus_of_variations_discrete}
\text{MSE}(h, x) = \frac{1}{m}\sum_{i=1}^m (h(x_i) - y_i)^2
\end{equation}
The output of a functional depends on not only the \emph{integrand/summand} $f(\cdot)$, but also the \emph{interval} (i.e., the data) for integration/summation, which theoretically explains why the training process is heavily data-dependent, and deficiencies in data themselves and data usage will impact learning-based algorithms. 
Although based on the \emph{Universal Approximation Theorem} \cite{csaji2001approximation}, deep networks can approximate the ``infinite-dimensional vector" $h(\cdot)$ with sufficient computation resource, they cannot resolve issues from data.   

\subsection{The Point-function Duality}

\begin{figure}
	\centering
    \includegraphics[width=.3\textwidth]{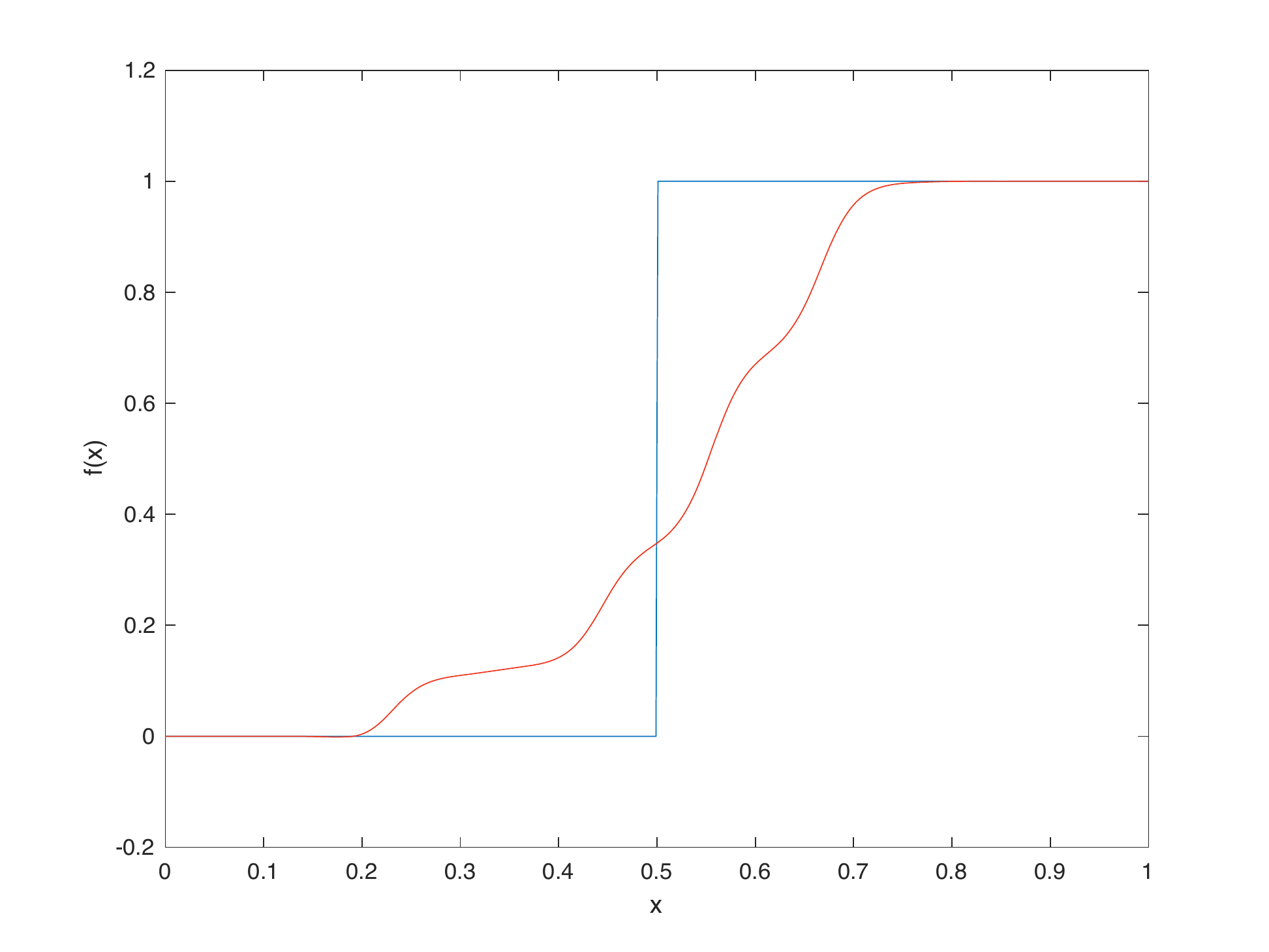}
    \caption{\label{fig:sample_effect} Training samples affect the geometry of the approximated space. Blue: 40,000 from [0.4, 0.6); orange: 20,000 from [0, 0.1) and 20,000 from [0.9, 1).}
\end{figure}

We introduce the \emph{point-function duality} of target functions to emphasize that the training process is \textbf{zero-constrained}. From the point perspective, $f(\cdot)$, as a point in the domain of loss functional $L(f)$, cannot determine what its actual domain and range are. From the function perspective, the topological property of its domain and range matters, especially if we apply $f:X\rightarrow Y$ to other inputs. For example, a regression algorithm fits the curve by minimizing square errors. The obtained function, however, different from \emph{admissible functions} in calculus of variations, sets no restrictions on the value of its inputs and outputs. 
In some cases, the target function might be locally valid around training data. As equation (\ref{eq:calculus_of_variations_discrete}) suggests, the output of the loss function depends on input data $\{x\}$ as well. If the function is trained using data within a certain neighborhood, dataset bias will occur. 
Figure \ref{fig:sample_effect} shows the curve fitting results of function \ref{eq:simple_step_function} with different training data. 
\begin{equation} \label{eq:simple_step_function}
f(x)=\left\{
\begin{aligned}
0, &  & x \leq 0.5 \\
1, &  & x > 0.5
\end{aligned}
\right.
\end{equation}
The shape of learned curves and boundaries is determined by training data and sometimes cannot be controlled directly by learning algorithms. Similarly, the decision boundaries for image classifiers will shift as training samples differ. Therefore, instead of resulting from insufficient data, dataset bias is an \textbf{intrinsic} property due to the point-function duality of target functions. 
\subsection{The Logic of Neural Networks}
\epigraph {When you have eliminated all which is impossible, then whatever remains, however improbable, must be the truth.} %
																											{Sherlock Holmes}
We claim that uncertainties in neural networks might be related to a flaw in current data usage, that is, \emph{limited options}. 
Let's put ourselves in the shoes of neural networks and rethink the reason for adversarial examples. Although output labels change dramatically after perturbation, networks may not agree that they are fooled as they still predict correct probabilities for most but not all of the remaining categories. In this sense, the networks are ``innocent'' as the ``winner-takes-all" rule of picking the label with the maximum likelihood is specified by network designers. We keep assuming casually that machine learning algorithms should always learn and follow the exact concepts (e.g., object classes) from human, but overlook the possibility that algorithms may require multiple classes to fully comprehend a human-level concept. Specifically, in an algorithm's perspective, it is possible that instances in one dataset are ``essentially" different from instances in another dataset, even though they correspond to the same concept according to humans. Therefore, we add a random-noise class for exceptions in section \ref{subsec:experiment_random-noise} and more equivalence classes with a tree structure in the last experiment of section \ref{subsec:experiment_quotient-space} to break these constraints.  

\begin{figure}
	\centering
    \captionsetup{justification=centering}
    \begin{subfigure}{.15\textwidth}
    	\includegraphics[width=\textwidth]{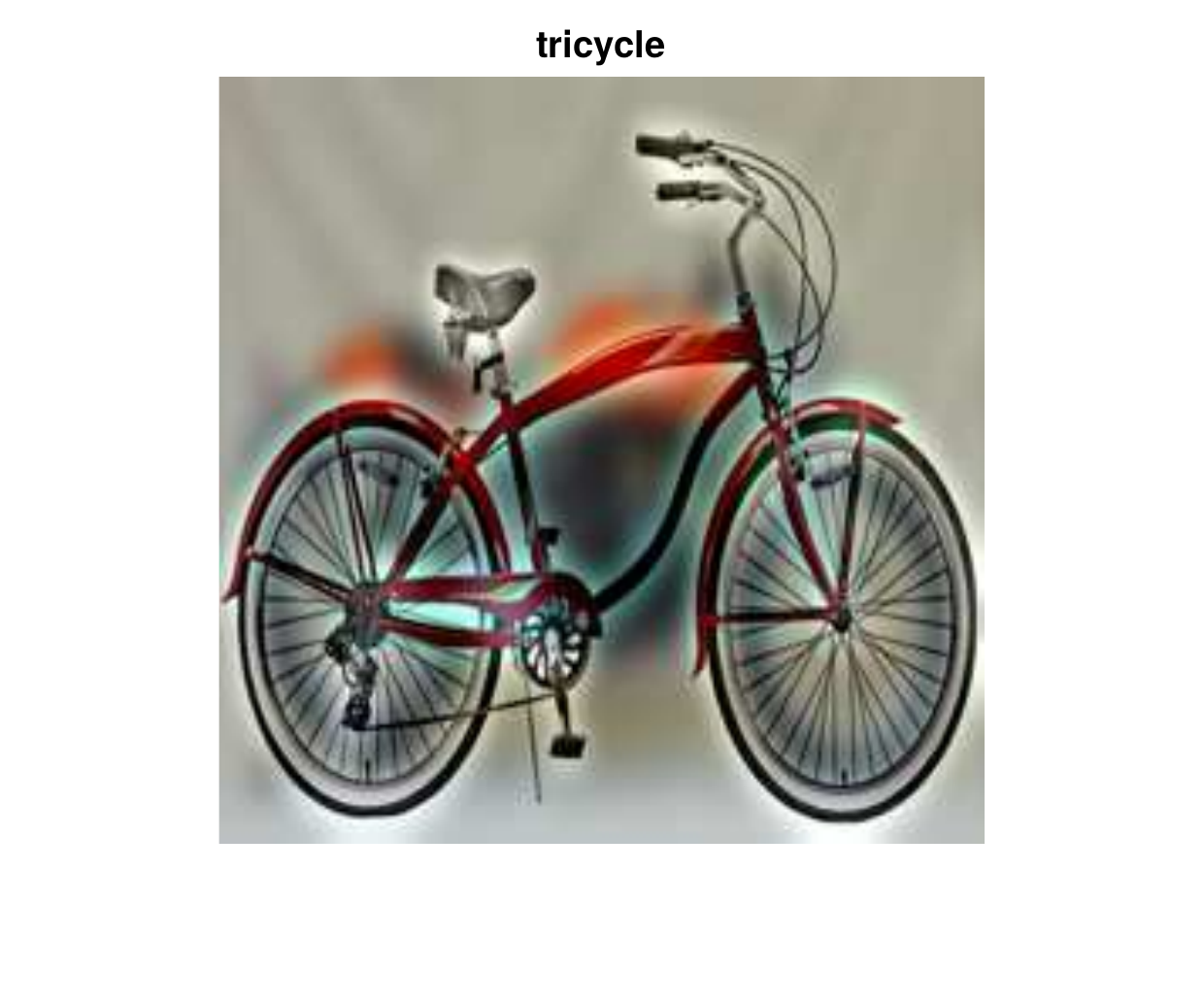}
        \centering \caption{bike + motorbike: tricycle}
    \end{subfigure}
    \begin{subfigure}{.15\textwidth}
    	\includegraphics[width=\textwidth]{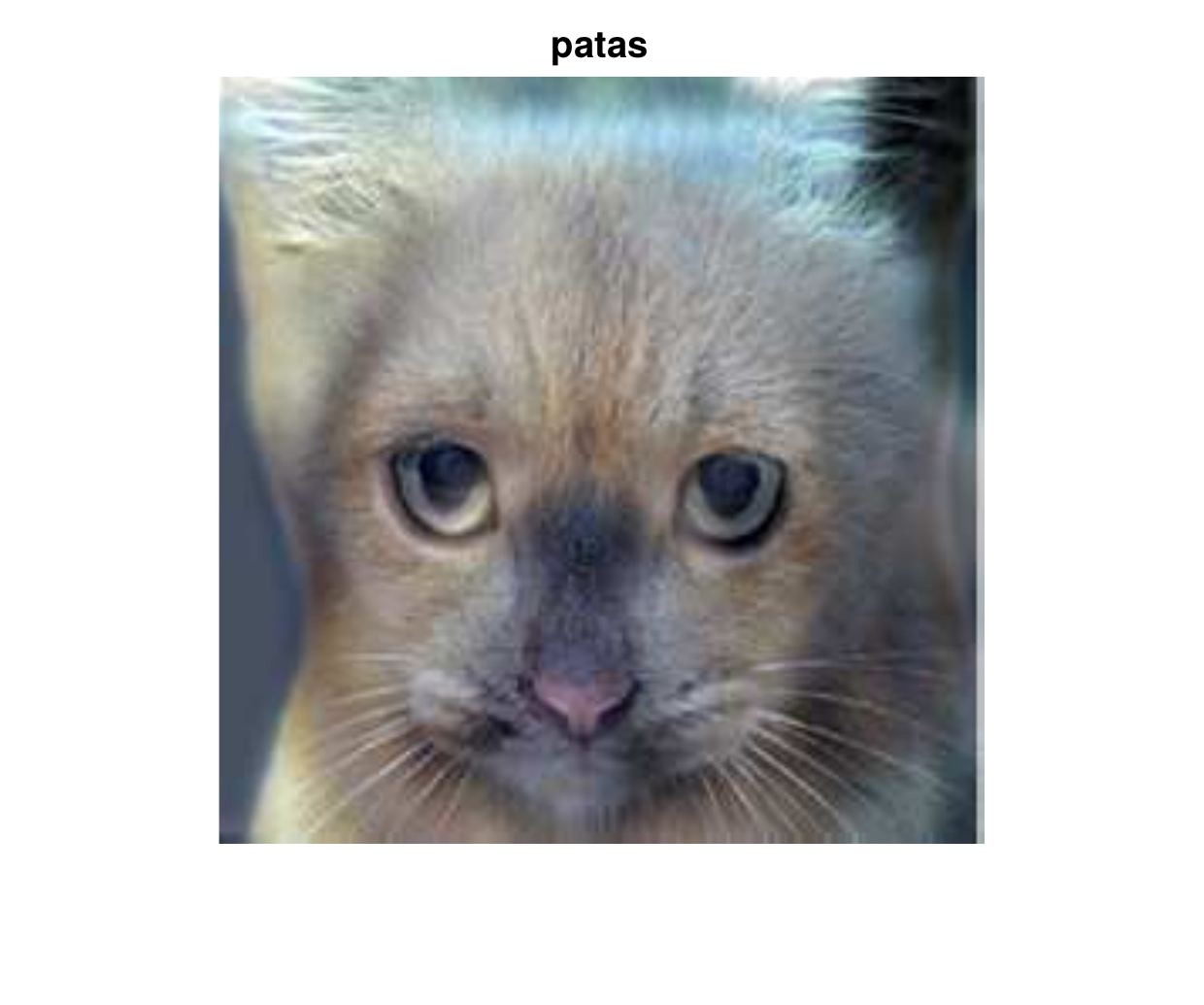}
        \centering \caption{cat + dog: patas}
    \end{subfigure}
    \begin{subfigure}{.15\textwidth}
    	\includegraphics[width=\textwidth]{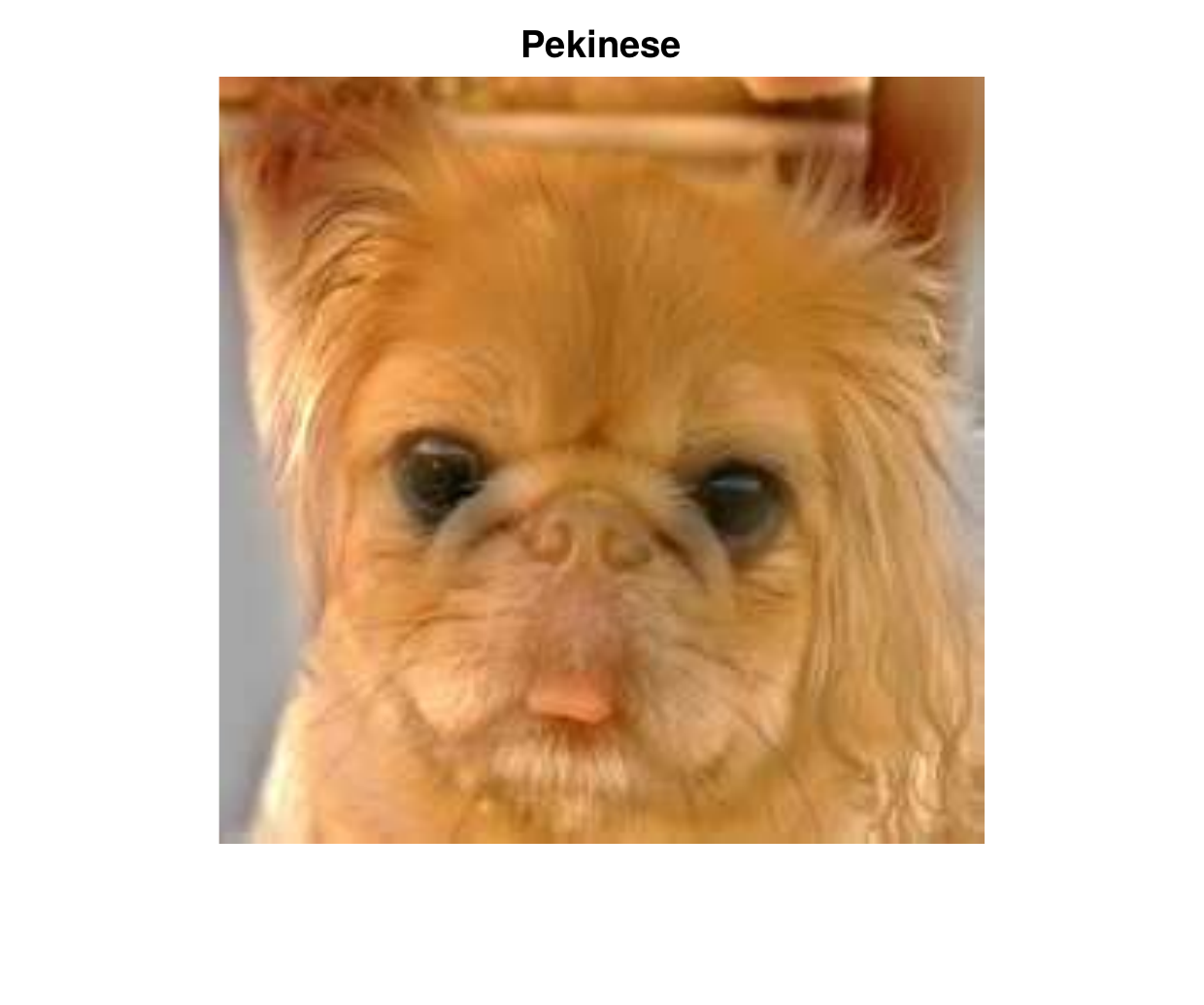}
        \centering \caption{dog + cat: Pekinese}
    \end{subfigure}
    \begin{subfigure}{.15\textwidth}
    	\includegraphics[width=\textwidth]{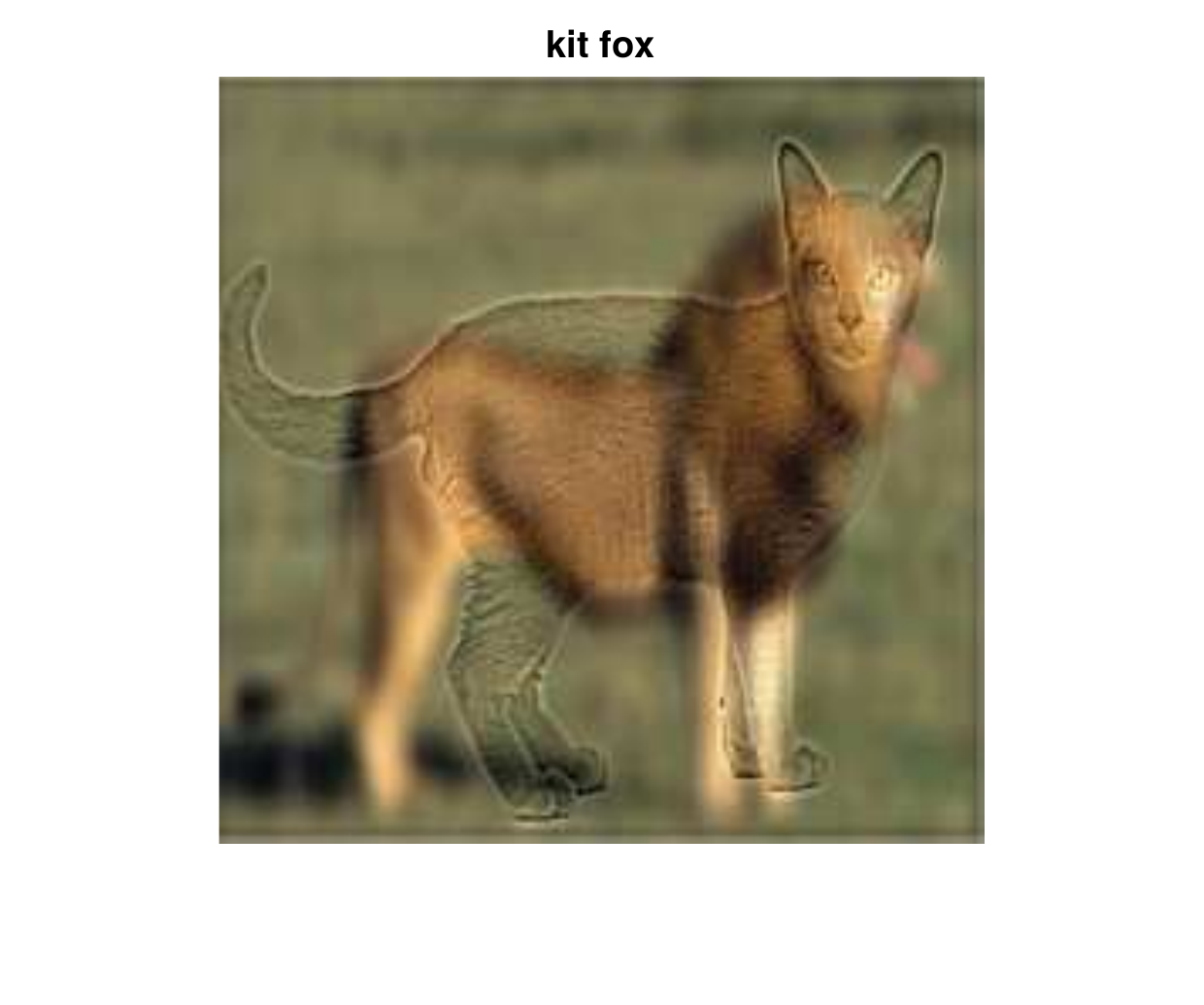}
        \centering \caption{lion + cat: \\kit fox}
    \end{subfigure}
    \begin{subfigure}{.15\textwidth}
    	\includegraphics[width=\textwidth]{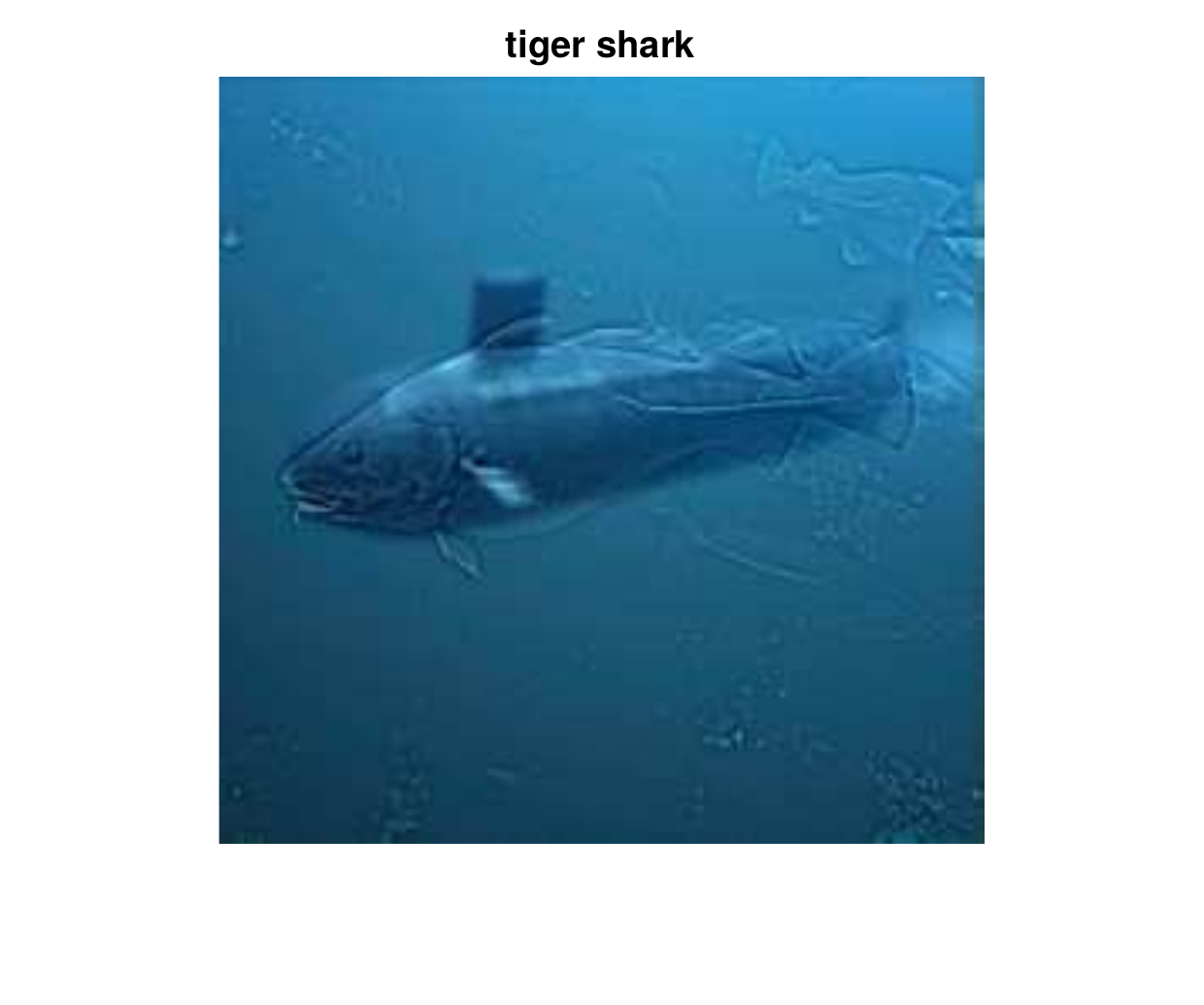}
        \centering \caption{fish + submarine: tiger shark}
    \end{subfigure}
    \begin{subfigure}{.15\textwidth}
    	\includegraphics[width=\textwidth]{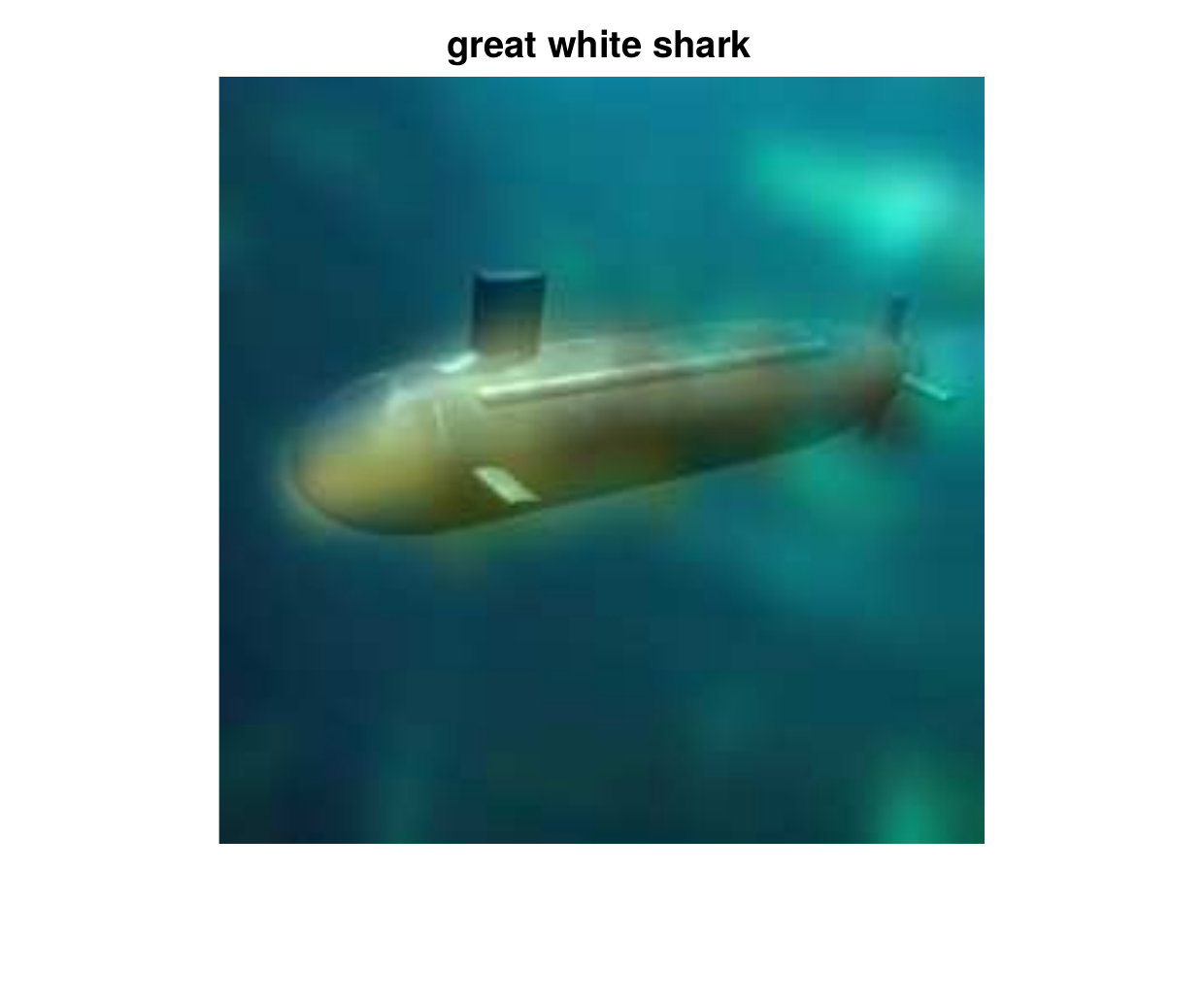}
        \centering \caption{fish + torpedo: great white shark}
    \end{subfigure}
    \caption{\label{fig:classification_hybrid} Classification for hybrid images.}
\end{figure}

We can analyze adversarial examples from the completely opposite perspective. During training, neural networks not only learn what an object ``is" but also capture what ``is not" an instance of a category. The mechanism of successful perturbations is to meet the criteria of ``is not." When a classifier eliminates all impossible categories with Sherlock's logic, it has to pick the remaining category as the final output. Unfortunately, no category for exceptions exists. Consequently, the classifier assumes that the target function is defined on all samples from the input space, which is rarely true. In addition, to classify the input space smoothly, the decision boundaries are deformed.
Figure \ref{fig:classification_hybrid} presents an interesting experiment in which the pretrained AlexNet \cite{krizhevsky2012imagenet} 
is asked to classify hybrid images \footnote{Selected from \url{https://www.cc.gatech.edu/classes/AY2016/cs4476_fall/results/proj1/}}
\cite{oliva2006hybrid} composed of high-frequency component of one image and low-frequency component of the other. Under current rules, the classifier has to choose one category from 1,000 available options; but neither ground-truth category is selected. 

\section{Models of Natural-image Spaces} \label{sec:image_space}
As the training process is heavily data-dependent, we deem that the characteristics of the input space are worth studying. Another motivation is that unlike decision trees, neural networks cannot classify even and odd numbers, which implies that the topological properties of the input space matter. 
In this section, we study the topological and geometric property of the natural-image spaces, as well as their consequences on computer-vision algorithms using neural networks. It is worth mentioning that both sections \ref{sec:philosophy} and \ref{sec:image_space} aim at providing motivation and intuition of this work, which guide the experiment design in section \ref{sec:experiments}. 
Related studies of the image space discussed in this section include image generation and translation with GANs \cite{goodfellow2016nips}.  
\subsection{Image Manifold in a Discrete Form}
The hypothesis of the continuous image manifold has been popular for decades. Before the deep-learning era, most classical computer-vision algorithms started from the continuous world by building features with physical meaning. Researchers have studied mappings from the image space to a target space, to solve tasks such as classification and segmentation. Because of recent developments in computation, estimating the reverse mapping, or even to traverse the image manifold \cite{zhu2016generative} is possible. However, there is no guarantee that all important topological and geometric properties in the continuous domain are preserved in such a discrete estimation unless we have a complete picture of the image space. In fact, although GANs can generate amazing images, the failure cases remain uncontrollable and sometimes difficult to explain. 
To match the discrete settings in GAN, we address natural image spaces in a completely discrete manner, hoping that we will gain a better understanding of current algorithms:
\begin{definition}{Discrete natural-image space:} 
Let $\mathds{Z}_{[l, u]}$ denote the set of integers from $l$ to $u$. Given an image $I$ with resolution $w\times h$ and $d$ channels, the image is considered as a point of the $\mathds{Z}_{[0, 255]}^{w\times h\times d}$- based natural image space $\mathds{I}^{w\times h\times d}$ if $I$ appears to be natural.
\end{definition}
The essence of the concept is to consider a space for each resolution as a ``chart" of the manifold, and these spaces are not necessarily smooth.   
\subsection{Properties of Natural-image Spaces}
In this subsection, we will describe what discrete natural-image spaces look like by listing important properties.

\emph{\bf{Sparsity}} and the \emph{\bf{scale-space effect}}: Obviously, the space is sparse as the probability that a random sample in $\mathds{Z}_{[0, 255]}^{w\times h\times d}$ falls on the corresponding natural-image space $\mathds{I}^{w\times h\times d}$ is small. 
When image resolution increases, the total number of cases increases exponentially. Similar to the effect in scale-space theory \cite{lindeberg1994scale}, humans are more sensitive to images with higher resolutions than to those with lower resolutions, meaning that flaws in high resolutions tend to be identified more easily. Hence, natural-image spaces are denser in lower resolutions because the quantity of valid natural images increases more slowly than that of possible cases. Such properties are consistent with results that show algorithms for image generation and translation appear to be more natural in images with relatively low resolutions. 

\emph{\bf{Connectivity:}}
Based on the sparsity, we infer that two images that belong to the same category are not always ``path-connected'' through adjacent grid points in $\mathds{Z}_{[0, 255]}^{w\times h\times d}$. 
We claim that a natural-image space can be regarded as a \emph{quotient space} consisting of numerous equivalence classes of path-connected images that can be derived from each other via non-destructive operations (e.g., using filters). In a classification task, one category contains multiple equivalence classes. These equivalence classes, however, are not necessarily connected.  
\begin{figure}
	\centering
	\begin{subfigure}{.20\textwidth}
    	\includegraphics[height=2.0cm]{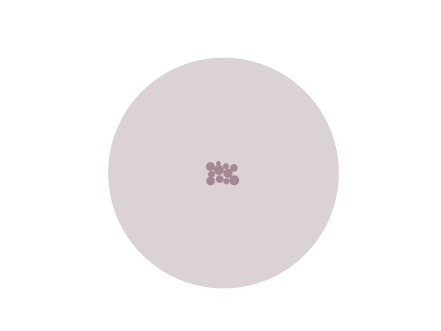}
        \centering \caption{Panoramic view}
    \end{subfigure}
    \begin{subfigure}{.20\textwidth}
    	\includegraphics[height=2.0cm]{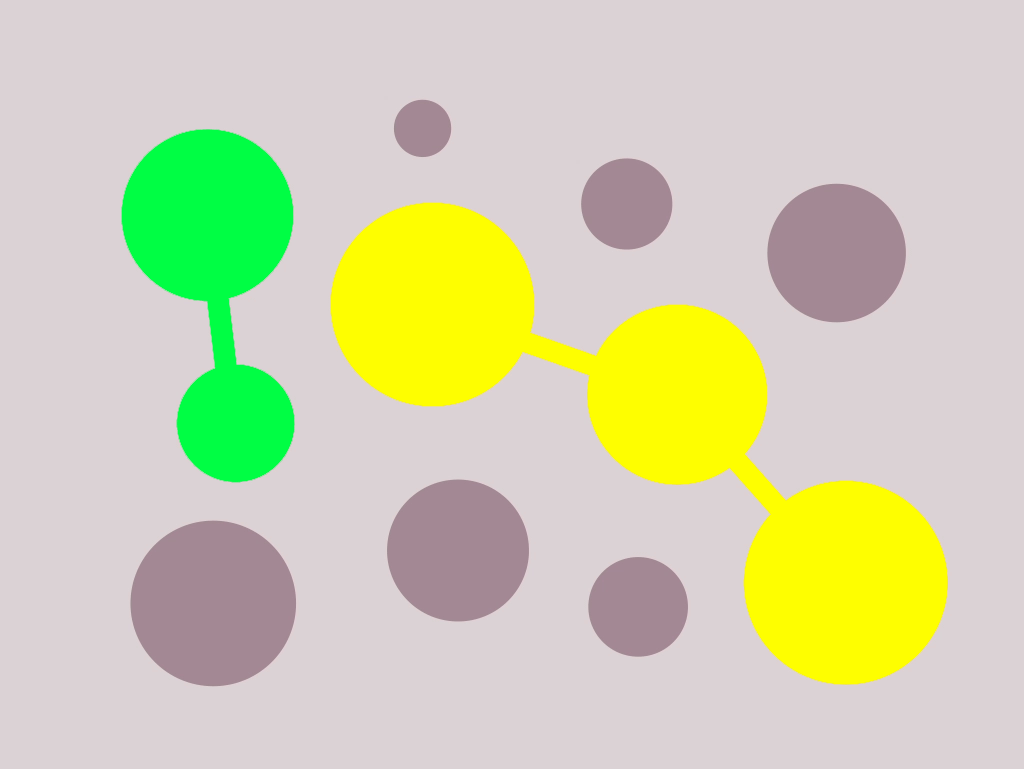}
        \centering \caption{Details }
    \end{subfigure}
    \caption{\label{fig:connectivity} Intuitive illustration on the connectivity of natural image space. Points represent natural images.}
\end{figure}
Figure \ref{fig:connectivity}
shows the conceptual model of a natural-image space in which each small darker disk represents a natural image, and images from the same equivalence class are connected. Under such settings, natural images do not share the same manifold; instead, each equivalence class has its own. 
\subsection{Approximation Challenge for Neural Networks} \label{subsec:challenge}
One gap exists between a labeled natural-image space (i.e., a quotient space) and the one approximated by a neural network, because the regions classified by the network tend to be connected \cite{fawzi2018empirical}. We claim that such a gap is one of the key reasons for uncertainties in neural networks and provide more experimental details in section \ref{subsec:experiment_quotient-space}. Intuitively, neural networks excel at drawing smooth decision boundaries in the input space,  whereas they lack the ability to slice the space and glue separate pieces together. To connect separate regions, a network needs to deform the decision boundary, creating somewhat arbitrary bridges that pass through uncertain regions in between. In contrast, decision trees can easily handle separated equivalent classes, which explains why they understand even and odd numbers better than neural networks.  
\subsection{Reinterpretation of Existing Work}
At the end of this section, we explain existing work from the viewpoint of this paper without proof. The essence of the explanation is to refresh our understanding towards experiment results from a unified yet nonexclusive perspective. The following opinions are provided as alternative and supplementary rather than a correction of the original claims.   
While image translation can be interpreted as finding a mapping between two points in a natural-image space, image generation seeks the reverse mapping from a latent space to a natural-image space. The major difficulty lies in the need for \textbf{designing an appropriate loss functional that indirectly regulates the performance of its extremum} (i.e., the target function). Effective approaches are found in recent studies such as those in \cite{isola2016image} and \cite{zhu2017unpaired}. 
Specifically, $L_1$ norm forces the image to be sufficiently sharp, which prevents the output from leaving a natural-image space. Strict rules for coloring adopted by the facade dataset \cite{tylevcek2013spatial} limits the cases of outputs, leading to a denser output space. 
Cycle loss manages to ensure the bidirectional property of the mapping, so that it obtains a more stable bijection. 
In \cite{karras2017progressive}, the network first locates natural-image clusters at lower resolutions, which is comparatively easier as the space is denser. With these coarse locations, outputs in higher resolutions are more likely to stay on natural-image spaces. 
The network then grows the scale progressively to refine outputs for better details.
In \cite{zheng2016improving}, stability-loss guarantees that the target function is locally consistent within the neighborhood of each image, providing buffer zones between images and decision boundaries. The authors of \cite{metzen2017detecting} proposed a subnetwork that distinguishes genuine data from perturbed ones, shrinking the natural-image spaces for the subsequent classifier. 
By assigning weights to training data, RetinaNet \cite{lin2017focal} detects objects more accurately at a fast speed.

\section{Experiments} \label{sec:experiments}
We present experimental results that support our statements on the impact of training samples and the benefits of utilizing the topological and geometric properties of natural-image spaces. As classification is the basis for advanced tasks, we set up CIFAR-10 \cite{krizhevsky2009learning} image classification experiments with a neural network adapted from the example code provided by MATLAB \cite{matlab2017cifardemo} and Keras \cite{keras2018cifardemo}. To demonstrate the effect of training samples, we design the experiments in a controlled manner and follow default configurations on the neural networks. For each experiment, we obtain final results by averaging results of 50 executions with fixed training data. Table \ref{tab:metric} describes the evaluation metric. We assume that the probability of each category is negatively correlated with the distance from the sample to the centroid of that category.
\begin{table}
	\caption{\label{tab:metric} Evaluation metrics used in experiments.}
	\centering
    \small
    \begin{tabular}{ cp{6.5cm} } 
    	\hline
         Metric & Explanation \\
         \hline
         $A$ & Average test accuracy \\
         $\sigma_A$ & Standard deviation of test accuracy \\
         $P_c$ & Average confidence of correctly classified samples \\ 
         $P_i$ & Average illusiveness (i.e., confidence) of misclassified samples \\ 
         $P_g$ & Average probability of the ground-truth category for misclassified samples \\ 
         $N$ &  Average number of test samples that are misclassified as ``noise'' (per 10,000) \\
        \hline
    \end{tabular}
\end{table}
\subsection{Impact of Training Samples} \label{subsec:impact_training-samples}
After training, we categorized the training samples into two classes: those that were correctly classified by the classifier and those that were misclassified by the classifier. We further sorted correctly-classified samples according to their confidence and select two subgroups with relatively higher ($S_{\text{hc}}$) and lower confidences ($S_{\text{lc}}$), respectively.  Similarly, we further selected two subgroups with higher illusiveness ($S_{\text{hi}}$) and lower illusiveness ($S_{\text{li}}$) from the misclassified training samples. With all subgroups the same size, we then retrained the network with the selected subgroups to illustrate the impact of different training samples. Table \ref{tab:retrain_subgroups} shows the performance of the classifier after retraining with subgroups of the original training data.

\begin{table*}
	\caption{\label{tab:retrain_subgroups} Performance of the classifier retrained with subgroups of training data. $S_\text{c}^p$: the top $p$ percent of training samples selected with criteria c. The number of uniformly distributed noise samples are $5\times$ the number of legitimate samples. }
	\centering
    \scriptsize   
	\renewcommand{\arraystretch}{1.2} 
    \begin{tabular}{ cccccc|cccccccc } 
    	\hline
        Training set & $A$ & $\sigma_A$ & $P_c$ & $P_i$ & $P_g$ & Training set & $A$ & $\sigma_A$ & $P_c$ & $P_i$ & $P_g$ & $N$ \\
        \hline
         $S_{\text{lc}}^{.25} \cup S_{\text{li}}^{.25}$ & 27.12\% & 2.40\% & 26.56\% & 22.91\% & 14.65\% & 
         $S_{\text{lc}}^{.25} \cup S_{\text{li}}^{.25} \cup $ noise & 30.82\% & 2.11\% & 28.31\% & 24.14\% & 15.43\% & 2.1 \\
         
         $S_{\text{lc}}^{.25} \cup S_{\text{hi}}^{.25}$ & 19.26\% & 0.92\% & 24.68\% & 22.86\% & 13.67\% &
         $S_{\text{lc}}^{.25} \cup S_{\text{hi}}^{.25} \cup $ noise & 21.32\% & 0.91\% & 25.27\% & 23.40\% & 14.30\% & 1.08 \\
         
         $S_{\text{hc}}^{.25} \cup S_{\text{li}}^{.25}$ & 53.61\% & 1.23\% & 79.67\% & 58.00\% & 12.73\% &
         $S_{\text{hc}}^{.25} \cup S_{\text{li}}^{.25} \cup $ noise & 55.27\% & 0.86\% & 79.24\% & 58.60\% & 12.86\% & 5.36 \\
         
         $S_{\text{hc}}^{.25} \cup S_{\text{hi}}^{.25}$ & 51.49\% & 1.29\% & 66.57\% & 44.77\% & 14.57\% &
         $S_{\text{hc}}^{.25} \cup S_{\text{hi}}^{.25} \cup $ noise & 53.26\% & 1.06\% & 67.22\% & 44.81\% & 14.87\% & 2.78 \\ 
        \hline
         $S_{\text{lc}}^{.50} \cup S_{\text{li}}^{.50}$ & 52.16\% & 1.58\% & 45.73\% & 36.39\% & 18.61\% & 
         $S_{\text{lc}}^{.50} \cup S_{\text{li}}^{.50} \cup $ noise & 53.11\% & 1.36\% & 46.70\% & 37.26\% & 19.23\% & 0.9 \\        
         
         $S_{\text{lc}}^{.50} \cup S_{\text{hi}}^{.50}$ & 41.07\% & 1.18\% & 35.11\% & 30.86\% & 17.99\% & 
         $S_{\text{lc}}^{.50} \cup S_{\text{hi}}^{.50} \cup $ noise & 40.82\% & 1.10\% & 35.05\% & 31.23\% & 18.55\% & 0.68 \\
         
         $S_{\text{hc}}^{.50} \cup S_{\text{li}}^{.50}$ & 65.78\% & 0.58\% & 88.08\% & 70.23\% & 11.41\% & 
         $S_{\text{hc}}^{.50} \cup S_{\text{li}}^{.50} \cup $ noise & 66.37\% & 0.53\% & 88.56\% & 70.03\% & 11.61\% & 1.12 \\
         
         $S_{\text{hc}}^{.50} \cup S_{\text{hi}}^{.50}$ & 60.48\% & 0.67\% & 75.62\% & 50.50\% & 15.92\% & 
         $S_{\text{hc}}^{.50} \cup S_{\text{hi}}^{.50} \cup $ noise & 65.13\% & 0.65\% & 76.10\% & 50.47\% & 16.16\% & 0.72 \\
        \hline
         $S_{\text{lc}}^{.75} \cup S_{\text{li}}^{.75}$ & 68.76\% & 1.16\% & 74.25\% & 52.34\% & 18.36\% & 
         $S_{\text{lc}}^{.75} \cup S_{\text{li}}^{.75} \cup $ noise & 70.03\% & 0.81\% & 73.88\% & 53.26\% & 18.42\% & 0.14 \\
         
         $S_{\text{lc}}^{.75} \cup S_{\text{hi}}^{.75}$ & 60.55\% & 1.31\% & 58.33\% & 43.00\% & 20.33\% & 
         $S_{\text{lc}}^{.75} \cup S_{\text{hi}}^{.75} \cup $ noise & 61.95\% & 1.38\% & 54.95\% & 43.80\% & 20.69\% & 0.50 \\
         
         $S_{\text{hc}}^{.75} \cup S_{\text{ls}}^{.75}$ & 70.81\% & 0.31\% & 92.67\% & 76.09\% & 10.28\% & 
         $S_{\text{hc}}^{.75} \cup S_{\text{li}}^{.75} \cup $ noise & 71.05\% & 0.54\% & 92.43\% & 75.64\% & 10.54\% & 0.22 \\
         
         $S_{\text{hc}}^{.75} \cup S_{\text{hi}}^{.75}$ & 70.85\% & 0.47\% & 80.45\% & 55.76\% & 16.24\% & 
         $S_{\text{hc}}^{.75} \cup S_{\text{hi}}^{.75} \cup $ noise & 71.42\% & 0.55\% & 80.94\% & 55.75\% & 16.48\% & 0.34 \\ 
        \hline
    \end{tabular}
\end{table*}
We observe the following on Table \ref{tab:retrain_subgroups}. 
More training samples do not guarantee a higher accuracy.
High-confidence images are required for higher test accuracy, especially when a limited number of training samples are provided. 
Highly illusive images are misleading when the size of the training set is small; as the training set expands, however, such adverse images become valuable and lead to even higher accuracy than training with low-illusiveness ones. One possible explanation is that highly illusive images, as outliers, force the network to adjust, which lowers loss. In this sense, the highly illusive images contain higher entropy (more information) than low-illusiveness images after a certain number of iterations. 
Classifiers trained with $S_{\text{hc}}\cup S_{\text{li}}$ are determined (smaller $\sigma_A$) and confident (higher $P_c$, $P_i$) about what they predict, regardless of whether they are right or wrong. By contrast, classifiers trained with $S_{\text{lc}}\cup S_{\text{hi}}$ are relatively hesitant (larger $\sigma_A$) in that the average probabilities for the output category ($P_c$, $P_i$) are lower; moreover, even if the prediction is wrong, they still assign a certain probability on the ground-truth category ($P_g$).  
In summary, the results demonstrate that the performance of classifiers depends on the training data. In other words, dataset bias occurs even within the same dataset because it is an intrinsic property of the learning scheme. 

\subsection{Training with a Random-Noise Class} \label{subsec:experiment_random-noise}
Adding random noise samples to training data is the easiest way to change the topology of the input space. 
The use of random noise in training neural networks dates back to the 1990s. Researchers have injected noise into inputs and weights to improve the generalization \cite{matsuoka1992noise}, avoid local minima, and speed up backpropagation \cite{wang1999training}. A recent study \cite{zheng2016improving} used random perturbation to stabilize the networks. However, researchers have never treated random noise as independent training samples containing information that can be directly employed for training neural networks. In this work, we add an extra category of random noise as negative samples, retrain the network, and test the classifier on the original test set that does not contain samples of random noise. Surprisingly, with random noise, we slightly improve the test accuracy at almost no extra cost except slightly longer computation time. 

\begin{figure*}
	\centering
    \footnotesize
	\begin{subfigure}{.45\textwidth}
    	\centering
    	\includegraphics[height=2.5cm]{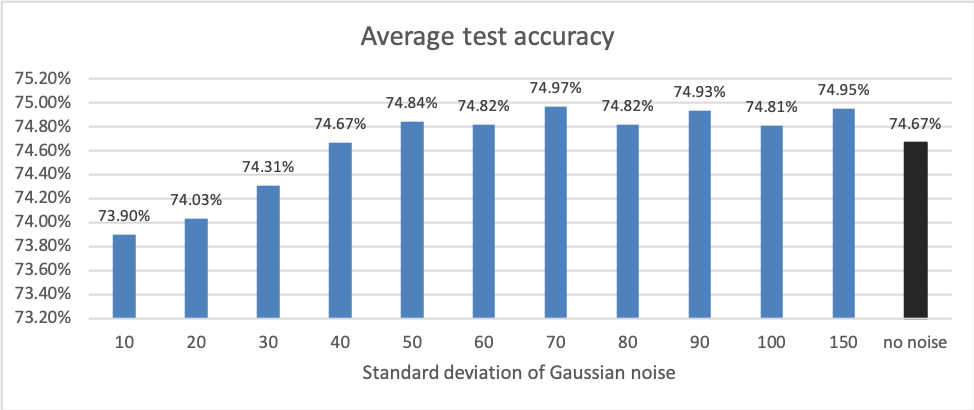}
        \captionsetup{justification=centering}
        \caption{\label{fig:Gaussian-std_vs_accuracy}  Average accuracy ($A$) vs. standard deviation (std). \\ $1\times$ noise samples centered at 127, scale 0}
    \end{subfigure}
	\begin{subfigure}{.45\textwidth}
    	\centering
    	\includegraphics[height=2.5cm]{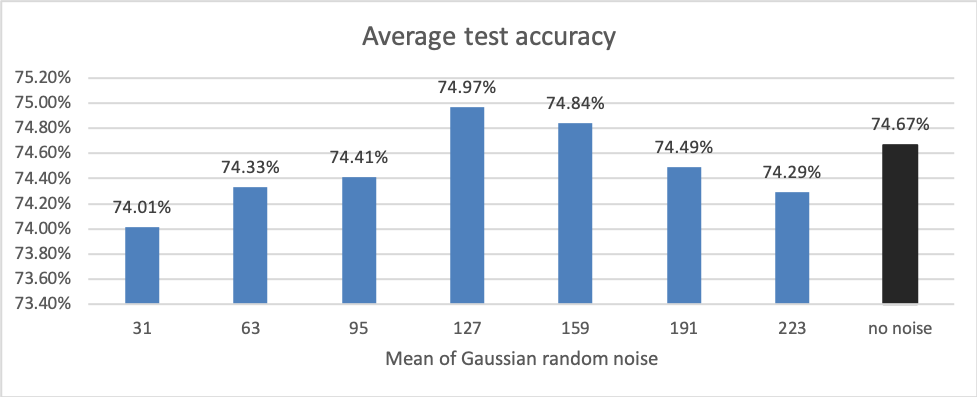}
        \captionsetup{justification=centering}
    	\caption{\label{fig:Gaussian-mean_vs_accuracy} Average accuracy ($A$) vs. mean. \\ $1\times$ noise samples with std 70, scale 0}
    \end{subfigure}  
    \begin{subfigure}{.45\textwidth}
    	\centering
    	\includegraphics[height=2.8cm]{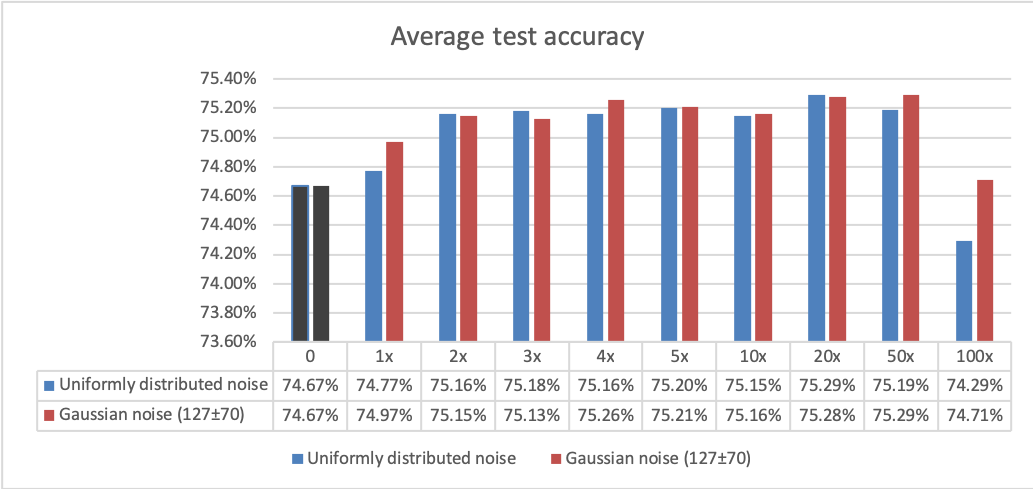}
        \captionsetup{justification=centering}
        \caption{\label{fig:noise-rate_vs_accuracy} Average accuracy ($A$) vs. rate. \\ Noise samples with mean 127, std 70, scale 0}
    \end{subfigure}
    \begin{subfigure}{.45\textwidth}
    	\centering
    	\includegraphics[height=2.8cm]{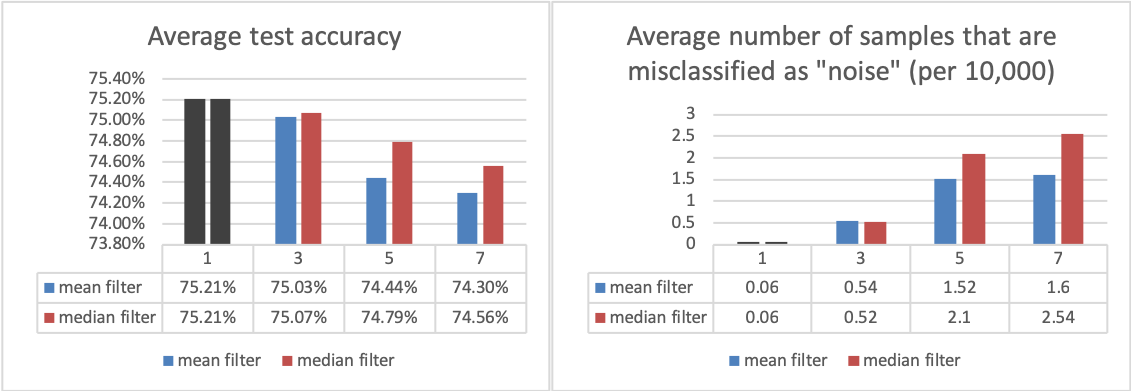}
        \captionsetup{justification=centering}
        \caption{\label{fig:Gaussian-scale_vs_performance} Average performance ($A$, $N$) vs. scale. \\ 5$\times$ noise samples with mean 127, std 70}
    \end{subfigure}
   	\caption{Controlled experiments demonstrating the effectiveness of each factor in random noise.}
\end{figure*}

We start testing with the most-common Gaussian random noise under four controlling parameters: mean, standard deviation, quantity, and scale (correlation among pixels). We design the experiments in a controlled manner to evaluate the effect of each factor. 
Figure \ref{fig:Gaussian-std_vs_accuracy} shows that Gaussian random noise improves test accuracy as long as the standard deviation is sufficiently large. As the standard deviation increases, Gaussian noise becomes more uniformly distributed. 

Figure \ref{fig:noise-rate_vs_accuracy} shows that test accuracy varies with different numbers of random-noise training samples. This finding suggests that random noise might contain a certain amount of information that could teach the classifier what ``is not'' a natural image. Such information, however, becomes saturated as the number of random-noise samples increases. With an excessive number (e.g., $100\times$) of noise samples, neural networks may become overwhelmed, spending much longer time in training but obtaining little enhancement. 

We further repeat the experiments in section \ref{subsec:impact_training-samples} with the additional category of uniformly distributed random-noise training samples, and report results at the right half of Table \ref{tab:retrain_subgroups}. Similar results with Gaussian random noise are provided in the supplementary materials. Comparing Tables \ref{tab:retrain_subgroups} and \ref{tab:retrain_subgroups}, we observe the following: Training classifiers with random noise tends to increase test accuracy, but such an effect is more obvious with fewer training samples. Despite the introduction of a new risk of ``misclassification as random noise'', it vanishes as the number of training images increases. In general, classifiers trained with random noise tend to be more determinant (higher $P_c$, $P_s$, lower $\sigma_A$); surprisingly, for misclassified samples, the probability of the ground-truth category also increases (higher $P_g$).  

The motivation of adding random noise is to change the topology of natural-image space by specifying the invalid zones that do not belong to the space. As a result, if we subtract the region of random noise from the input space of the classifier, then the input space is no longer ``dense" and has more holes. Geometrically, to avoid such holes, random noise tends to push the decision boundaries towards the centroid of each category. 
In addition, feeding neural networks with random-noise samples that are completely unnatural can help them learn what ``is not'' an object. 
However, as the scale of Gaussian noise increases, more pixels are correlated and samples become less unnatural. Therefore, the effect is weakened by Gaussian noise with larger correlations among pixels. 

At the end of this experiment, we report two cases that lead to interesting results. In one case, the test accuracy drops if we train the classifier with images of solid colors. 
The set of solid-color images can be interpreted as an embedded sub-hyperplane that 1) has the same or even lower dimension as decision boundaries, and 2) spreads across the entire input space. Classifying such sets in lower dimensions is difficult because their boundaries reside in even lower dimensions. For a better understanding, we may consider solving a curve-fitting problem with classification instead of regression.   
In the other case, test accuracy also drops in cases of Gaussian random noise with various means, standard deviations, and scales. Such mixed samples come from several separated clusters in the input space. Since neural networks group samples into path-connected regions \cite{fawzi2018empirical}, mixed noise becomes a more difficult case.  

Before we move on to the next experiment, it is worth reiterating that the major discovery here is not the marginal improvement on classification with a random-noise class. Instead, the existence of these unusual samples is much more inspiring, and they do not have to be random noise at all times. These samples used to be considered as ``off-the-manifold" and completely irrelevant to the classification task. After obtaining the experiment results, we look forward to seeking more hidden samples that may complement data augmentation, especially when training samples are insufficient.
\subsection{Training in the Natural-image Spaces} \label{subsec:experiment_quotient-space}
In addition to changing the topology of the labeled input space with an artificial class, we can also improve the classifier by slicing the connected regions learned by a network classifier to match the quotient-space model. This subsection continues the discussion on the quotient space in section \ref{subsec:challenge} and provides quantitative experiments that support our claim. Let us first recognize that a network with a given architecture and hyper-parameters has a limited, finite capability of approximating the quotient space. As a result of this, there will often be training samples that are incorrectly classified during training, sometimes even with very high confidence. Henceforth, we will refer to these samples as illusive samples. 
Higher-order statistics of the training process, including the number of times that a training sample has been correctly classified, indicate crucial characteristics of the learned space. We then conduct experiments based on the higher-order statistics collected during training.

\begin{figure}
\begin{center}
	\includegraphics[width=0.30\textwidth]{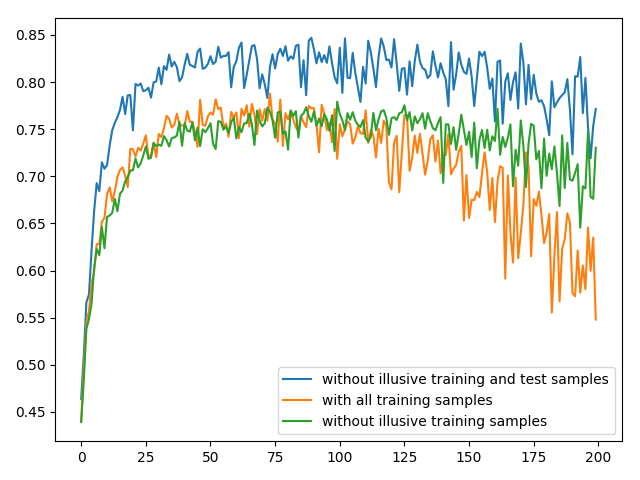}
\end{center}
   \caption{Training without consistently misclassified samples help prevent over-fitting.}
\label{fig:hard-sample_overfitting}
\end{figure}

We begin by retraining the network without these so-called illusive training samples that are consistently misclassified. As Figure \ref{fig:hard-sample_overfitting} shows, despite a slight drop in test accuracy during early epochs, over-fitting is alleviated. If we further remove test samples that are consistently misclassified (i.e., we ``cheat" a bit as our goal is studying the topological properties of the input space rather than achieving state-of-the-art accuracy), over-fitting is weakened even more. A correlation seems to exist between over-fitting and the illusive samples that cannot be mapped to the right equivalence classes. Moreover, the rationale behind dataset bias and over-fitting may essentially be the same; the difference is that over-fitting occurs at equivalence classes that are already observed whereas dataset bias happens at unseen locations in the input space. 

Following the statements in section \ref{subsec:challenge}, another consequence of removing illusive training samples is that classifiers sacrifice less the space in between to connect separated equivalence classes, which could potentially enhance the robustness against adversarial examples. We retrain the CIFAR-10 demo in CleverHans \cite{papernot2018cleverhans} and obtain results in Table \ref{tab:cleverhans} as expected. 
\begin{table}
	\caption{\label{tab:cleverhans} Test accuracy (in \%) from CleverHans. $n$: number of training epochs; $\epsilon$: max-norm eps; $A_{\text{leg}}$/$A_{\text{adv}}$: average test accuracy (20 runs) on legitimate/adversarial samples; $A^*_{\text{leg}}$/$A^*_{\text{adv}}$: average accuracy without illusive training samples.}
	\centering
    \scriptsize 
    \begin{tabular}{ p{.03cm}p{.03cm}p{.4cm}p{.4cm}p{.4cm}p{.4cm}|p{.03cm}p{.03cm}p{.4cm}p{.4cm}p{.4cm}p{.4cm}} 
    	\hline
         $n$ & $\epsilon$ & $A_{\text{leg}}$ & $A^*_{\text{leg}}$ & $A_{\text{adv}}$ & $A^*_{\text{adv}}$ & 
         $n$ & $\epsilon$ & $A_{\text{leg}}$ & $A^*_{\text{leg}}$ & $A_{\text{adv}}$ & $A^*_{\text{adv}}$ \\
         \hline
         6   & .3         & 78.90            & 78.88              & 10.84            & 11.74              &        
         50  & .3         & 90.39            & 88.81              & 10.37            & 10.40              \\
         
         6   & .1         & 78.73            & 78.22              & 9.34             & 11.31              &
         50  & .1         & 90.51            & 88.77              & 10.75            & 14.35              \\
         
         6   & .05        & 78.98            & 78.21              & 11.16            & 13.87              &
         50  & .05        & 90.53            & 88.91              & 13.28            & 21.88              \\ 
         
         6   & .01        & 78.84            & 78.73              & 45.47            & 48.39              &
         50  & .01        & 90.33            & 88.90              & 39.03            & 48.50              \\
        \hline
    \end{tabular}
\end{table}
Similarly to the results from other sophisticated adversarial training techniques that require carefully crafted adversarial loss, we achieve higher accuracy on adversarial examples with a slightly lower accuracy on legitimate samples, simply by training the classifier without illusive samples that are consistently misclassified. 

\begin{figure*}
	\centering
    \begin{subfigure}{.38\textwidth}
    	\centering
    	\includegraphics[height=3.0cm]{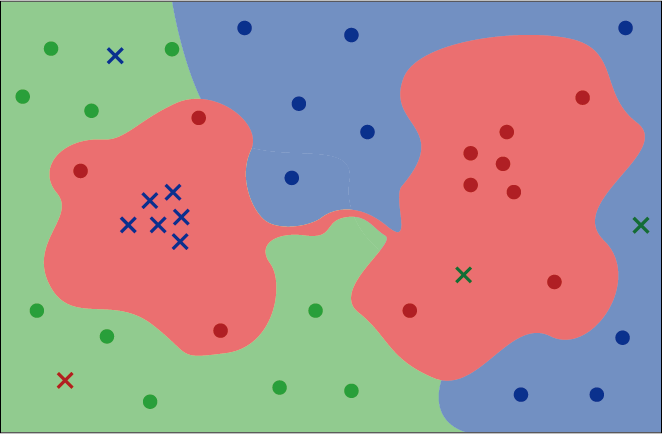}
        \captionsetup{justification=centering}
        \caption{\label{fig:connected_regions}Connected regions classified by a neural network}
    \end{subfigure}
    \begin{subfigure}{.38\textwidth}
    	\centering
    	\includegraphics[height=3.0cm]{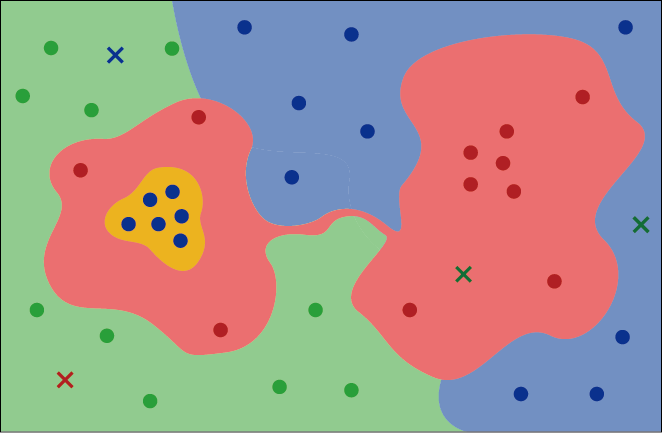}
        \captionsetup{justification=centering}
        \caption{\label{fig:quotient_space}The actual quotient space of the images}
    \end{subfigure}
    \begin{subfigure}{.20\textwidth}
    	\begin{tabular}{cp{1.5cm}} 
        	\hline
             Category & Equivalence class(es) \\
             \hline
             red & \textcolor{RedOrange}{$\medbullet$} \\
             green & \textcolor{LimeGreen}{$\medbullet$} \\ 
             blue & \textcolor{CadetBlue}{$\medbullet$} \textcolor{Dandelion}{$\medbullet$} \\
            \hline
        \end{tabular}
        \captionsetup{justification=centering}
        \caption{\label{tab:equivalence_classes}Look-up table for equivalence classes}
    \end{subfigure}
    \caption{\label{fig:quotient_space_strategy}Illustration on the potential of improving classification with the quotient-space model}
\end{figure*}
Our last experiment is a pure proof-of-concept, in which we assume regardlessly that prior knowledge is available of whether a test sample is illusive. During training, we compute the cumulative confusion matrix based on training samples. If an illusive training sample is consistently misclassified and appears to be one of the top confusion (i.e., large numbers in the cumulative confusion matrix), we relabel the sample with a new category. From the perspective of the quotient-space model, the network successfully learns that the illusive sample belongs to a particular equivalence class, but failed to connect the equivalence class with others in the same category due to limited capability of approximating the image space. Thus, as shown in Figure \ref{fig:quotient_space_strategy}, we slice the equivalence class with a new label, train another separate classifier based on the illusive samples, and glue the equivalence class to easy ones with a look-up table (or dictionary/decision tree). When illusiveness of test samples is available as prior knowledge, we can select which classifier to apply at test time. According to test results, the proposed strategy can potentially increase test accuracy by 3\% without requiring additional hardware. In addition, we can train networks with larger learning rate (i.e., faster) because over-fitting is weakened by complying with the image space approximated by the network.

\section{Discussion} \label{sec:discussion}
The primary contribution of this paper is the intrinsic and unified explanation of uncertainties in neural networks that are applied to images. All experiments are designed by following the explanation. In contrast, the results of the experiments are just the tip of the iceberg.
As the training process is data-dependent and zero-constrained, the learned classifier may suffer from dataset bias. 
Several key topological and geometric properties of smooth manifolds are missing in the discrete natural-image spaces adopted by neural networks, which we believe is part of the reason for adversarial examples as well as baffling outcomes from image translation and generation. 
Under current settings, the networks take it for granted that natural-image spaces are dense and simply return the most-likely output from limited options. Restrictions on the input domain and output range of neural networks are lacking. Experiment results suggest that adding an additional category of negative training data to handle exceptions might be helpful. 
Other results demonstrate that training networks according to the properties of the input space suggested from higher-order statistics has the potential to mitigate over-fitting, enhance robustness against adversarial examples, and improve the test accuracy for a given network structure without much more computation resource. 
These phenomena lead to the following questions: What data composition is better for training more robust classifiers and detectors? How do we measure the value or impact of training samples? What other characteristics of the input space can guide the training process?

We hope the paper 
will stimulate discussion in the community regarding the intrinsic properties of the input space to which deep networks are applied. 
Open problems include the entropy of training samples, features of the decision boundaries, equivalent relation among images, and better representation of the image space. Understanding the topological and geometric properties of natural-image spaces with a rigorous model will help us interpret the performance of state-of-the-art deep neural networks. Moreover, it may provide a more comprehensive understanding of the theoretical basis for deep neural networks.
In practice, we may enhance the performance of neural networks by improving the quality of training samples or altering how we use data.

\section{Acknowledgement}
We thank Dr. James Hays for the inspiring discussion in CS 7476 and the chance to present part of the work in class.

{\small
\bibliographystyle{ieeetr}
\bibliography{main}
}

\end{document}